\documentclass[lettersize,journal]{IEEEtran}
\usepackage{amsmath,amsfonts}

\usepackage{cite}
\usepackage{amsmath,amssymb,amsfonts}
\usepackage{algorithm}
\usepackage[noend]{algpseudocode}
\usepackage{graphicx}
\usepackage{textcomp}
\usepackage[table]{xcolor}
\definecolor{mygray}{gray}{0.95}
\definecolor{mygreen}{rgb}{0.78, 0.93, 0.8}
\definecolor{myred}{rgb}{1.0, 0.68, 0.69}
\definecolor{forestgreen}{RGB}{34,139,34}
\definecolor{lightblue}{RGB}{173,216,230}
\usepackage{xcolor}
\definecolor{lightgray}{gray}{0.9} 
\usepackage{url}
\usepackage{algpseudocode}
\usepackage{hyperref}
\usepackage{listings}
\def\BibTeX{{\rm B\kern-.05em{\sc i\kern-.025em b}\kern-.08em
    T\kern-.1667em\lower.7ex\hbox{E}\kern-.125emX}}

\definecolor{light-gray}{gray}{0.95}
\newcommand{\code}[1]{\setlength{\fboxsep}{0pt}\colorbox{light-gray}{\texttt{#1}}}

\lstdefinestyle{codestyle}{
  basicstyle=\ttfamily\small,
  keywordstyle=\color{black},
  stringstyle=\color{black},
  commentstyle=\color{black},
  backgroundcolor=\color{gray!10},
  frame=single,
  frameround=tttt,
  rulecolor=\color{black},
  breaklines=false,
  showstringspaces=false,
  upquote=true,
}

\newcommand{\codetable}[1]{\setlength{\fboxsep}{0pt}{\texttt{#1}}}

\lstdefinestyle{codestyle}{
  basicstyle=\ttfamily\small,
  keywordstyle=\color{black},
  stringstyle=\color{black},
  commentstyle=\color{black},
  backgroundcolor=\color{white},
  frame=single,
  frameround=tttt,
  rulecolor=\color{black},
  breaklines=false,
  showstringspaces=false,
  upquote=true,
}

\begin{document}

\title{SlicerROS2: A research and development module for image-guided robotic interventions}

\author{Laura Connolly$^{1,2}$, \emph{Student Member}, IEEE, Aravind S. Kumar$^{2}$, Kapi Ketan Mehta$^{2}$, Lidia Al-Zogbi$^{2}$,  \\ Peter Kazanzides$^{2}$, \emph{Member}, IEEE, Parvin Mousavi$^{1}$, \emph{Senior Member}, IEEE, 
Gabor Fichtinger$^{1}$, \emph{Fellow}, IEEE, Axel Krieger$^{2}$, \emph{Senior member}, IEEE, Junichi Tokuda$^{3}$, \emph{Member}, IEEE, Russell H. Taylor$^{2}$, \emph{Life Fellow}, IEEE, \\ Simon Leonard$^{2}$, \emph{Member}, IEEE, Anton Deguet$^{2}$, \emph{Member}, IEEE 
\thanks{$^{1}$ Queen's University, Kingston, ON, Canada. Corresponding author email: \emph{laura.connolly@queensu.ca},
        $^{2}$ Johns Hopkins University, Baltimore, MD, United States,
        $^{3}$ Brigham and Women's Hospital, Boston, MA, United States. This work is supported by National Institutes of Health (NIH) R01EB020667 (MPI: Tokuda, Krieger, Fuge, Leonard) and was completed as part of the Mitacs Globalink Research Award (GRA) program. We would also like to acknowledge the National Sciences and Engineering Research Council of Canada (NSERC) and the Canadian Institutes of Health Research (CIHR). Laura Connolly is supported by an NSERC Canada Graduate Scholarship-Doctoral (CGS-D) award, a Michael Smith Foreign Study Supplement (MS-FSS) and a Walter C. Sumner Memorial Fellowship. G. Fichtinger is a Canada Research Chair in Computer Integrated Surgery, Tier 1. Russell H. Taylor and Peter Kazanzides are supported by Johns Hopkins University internal funds.}
}

\markboth{Journal of \LaTeX\ Class Files,~Vol.~14, No.~8, August~2021}%
{Shell \MakeLowercase{\textit{et al.}}: A Sample Article Using IEEEtran.cls for IEEE Journals}

\IEEEspecialpapernotice{%
  {\small
  © 2024 IEEE. Personal use of this material is permitted. Permission from IEEE must be obtained for all other uses, in any current or future media, including reprinting/republishing this material for advertising or promotional purposes, 
  creating new collective works, for resale or redistribution to servers or lists, or reuse of any copyrighted component of this work in other works. This is the author’s accepted manuscript of an article published in \emph{IEEE Transactions on Medical Robotics and Bionics}. The final version of record is available at \url{https://doi.org/10.1109/TMRB.2024.3464683}.
  }%
}

\maketitle

\begin{abstract}
Image-guided robotic interventions involve the use of medical imaging in tandem with robotics. SlicerROS2 is a software module that combines 3D Slicer and robot operating system (ROS) in pursuit of a standard integration approach for medical robotics research. The first release of SlicerROS2 demonstrated the feasibility of using the C++ API from 3D Slicer and ROS to load and visualize robots in real time. Since this initial release, we've rewritten and redesigned the module to offer greater modularity, access to low-level features, access to 3D Slicer's Python API, and better data transfer protocols. In this paper, we introduce this new design as well as four applications that leverage the core functionalities of SlicerROS2 in realistic image-guided robotics scenarios. 
\end{abstract}

\begin{IEEEkeywords}
Image-guidance, robotics, 3D Slicer, ROS2, research prototyping, virtual fixtures, motion planning.
\end{IEEEkeywords}

\section{Introduction}
\IEEEPARstart{M}{edical} robotics is an evolving and rapidly growing research field with the potential to transform standard clinical practice. It is possible that robots will one day transcend human capabilities while offering higher efficiency, lower costs, improved training outcomes and better safety \cite{taylor2016medical}. The advancement of image-guided robotics in particular, which are systems that rely on both medical imaging and robotics, is critical for achieving this potential. This is because image-guided robots can be used to fuse preoperative and intraoperative realities \cite{fichtinger2022image}.

There are several combinations of imaging modalities and robotic systems have been explored in this capacity. For example, the SpineBot uses computed tomography (CT) imaging to help define the trajectory of pedicle screws, and robotics to guide the surgeon through those trajectories \cite{kochanski2019image}, \cite{chung2004development}. Another example is the MrBot, which was designed to help perform percutaneous needle interventions within the confines of a magnetic resonance imaging (MRI) scanner \cite{stoianovici2007mri}. Similarly, the Artemis robot was designed to facilitate transrectal prostate biopsy under ultrasound guidance with MRI fusion  \cite{sonn2014target}. These are just a few examples of the numerous procedures and therapies where the use of image-guidance in tandem with robotics has been investigated. More recently, advancements in image-guided robotics have enabled: navigation of catheters into blood vessels with magnetic continuum devices \cite{dreyfus2024dexterous}, autonomous needle steering for lung biopsy \cite{kuntz2023autonomous} and teleoperated neurovascular interventions \cite{kim2022telerobotic}.

Despite this extensive investigation, there are only a few areas of intervention where image-guided robotic systems that have achieved widespread adoption and financial commercial success such as robotic bronchoscopy, radiation oncology and neurosurgery \cite{fichtinger2022image} \cite{graetzel2019robotic}. One contributor to this slow growth and adoption is the lack of a common integration approach. For any image-guided robotic system, integration of the imaging modality and the robot is the most important factor for usability. However, several companies and research systems take their own unique approach to integration. This results in device-specific software, expensive research licenses, incompatible communication protocols, and overall, a high barrier to entry to develop such systems. Considering these challenges and their potential threat to continued development, it is imperative to provide a common integration scheme for image-guided robotics. We hypothesize that this will prevent re-engineering and promote reproducibility across different clinical applications.

\begin{figure*} [t]
    \centering
    \includegraphics[width=15 cm]{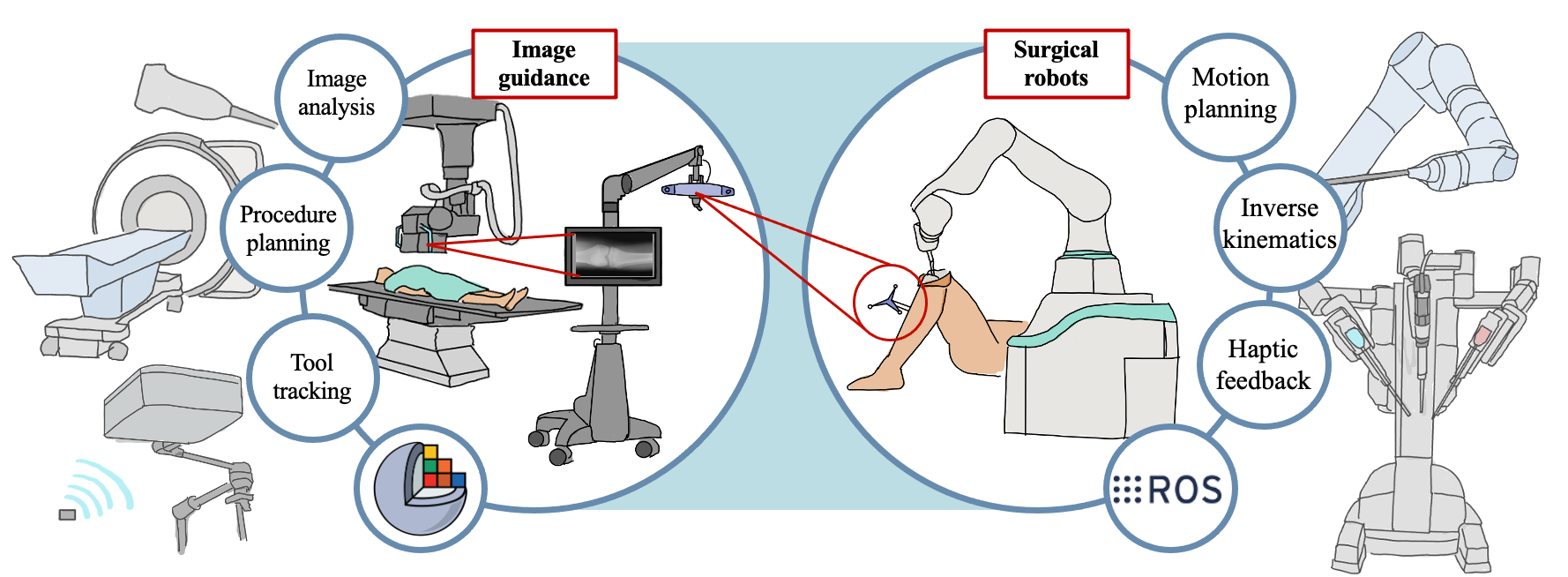}
    \caption{Overview of division between image-guidance and robotics research tools and standard algorithms / workflow steps. Left side shows the devices and tools that are typically used for IGT systems, as well as some of the features available in 3D Slicer (image analysis, procedure planning, tool tracking). Right side shows examples of surgical robots as well as the features available in ROS (haptic feedback, inverse kinematics, motion planning). An example robotic IGT system (robot guided knee surgery) is highlighted in the center of the figure. }
    \label{fig:igtrobot}
\end{figure*}

From a development perspective, image-guided therapy (IGT) platforms and medical robotics platforms are often separated. In the realm of IGT, the open-source medical imaging platform, 3D Slicer is the most commonly used research platform \cite{fedorov20123d} \cite{ungi2016open}. With over one million downloads and an active research and support community, 3D Slicer is used for segmentation, virtual reality, image analysis, artificial intelligence, and several other applications \cite{kapur2016increasing}. Several research platforms have been enabled by or added to 3D Slicer such as SlicerIGT \cite{ungi2016open}, SlicerVR \cite{pinter2020slicervr}, MONAI label \cite{diaz2022monai}, and Total Segmentator \cite{wasserthal2022totalsegmentator}. There are also open-source platforms like OpenIGTLink and the PLUS toolkit that allow users to interface commercial hardware with 3D Slicer to build complete IGT systems \cite{tokuda2009openigtlink} \cite{lasso2014plus}. As a result of these efforts, 3D Slicer is considered the de-facto open source software for developing navigated, image-guided interventions.

In the field of medical robotics, robot operating system (ROS), an open-source middleware designed to support robotics development, is the predominant framework. ROS is a modular development framework that provides tools for autonomous navigation, simulation, visualization, and control \cite{macenski2022robot}. Like 3D Slicer, ROS has a very active research community that is constantly contributing to the platform. For medical robotics specifically, several research tools like the Computer Integrated Surgical Systems Surgical Assistant Workstation (CISST-SAW) libraries \cite{deguet2008cisst} and the Asynchronous Multi-Body Framework (AMBF) \cite{munawar2019real} support ROS. The da Vinci Research Kit (dVRK), a popular open-source medical robot  \cite{kazanzides-chen-etal-icra-2014}, also supports ROS for software development. Furthermore, many commercially available robots provide a ROS interface off the shelf.

In an effort to pursue a common integration scheme for image-guided robotics research, we decided to bridge these two ecosystems. As evidenced by the numerous published papers that employ both 3D Slicer and ROS (71 papers available on Google Scholar using keywords ``3D Slicer" AND ``robot operating system"), there is also demand from the community for this integration. Previous attempts to bridge 3D Slicer and ROS such as the ROS-IGTL bridge \cite{tauscher2015openigtlink}, custom applications for specific robots \cite{connolly2021open} \cite{frank2017ros}, and our initial offering of SlicerROS2 \cite{connolly2022bridging} have fallen short of meeting all of the needs of a common integration tool. These needs include greater usability by providing access to low-level features, robust data transfer protocols that support commonly used message types, thorough documentation, and maintainability \cite{herz2019open}. We have since redesigned SlicerROS2 to further support image-guided robotics research considering these requirements. The details of this new design are described in the following sections.  The contributions of this paper are: 1) A newly designed research module for efficient data transfer between 3D Slicer and ROS 2 and 2) four relevant applications that demonstrate how it can be used for rapid research prototyping.

\section{Methodology}

\subsection{Building robotic IGT systems}

When developing robotic IGT systems, effectively communicating the patient's position to the robot is crucial. This communication is achieved with tracking via electromagnetic (EM) or optical sensors that are fixed to the patient and localized in the same coordinate space as the robot. Standard procedures and algorithms for tool calibration and system registration are typically applied to align these coordinate systems. Furthermore, if preoperative images are available, they are also aligned with the physical patient to assist in preoperative planning and real-time surgical guidance. The computational tools and visualization requirements for these workflow steps are freely available in 3D Slicer and have been extensively tested and integrated into existing IGT systems. Moreover, 3D Slicer supports various imaging and tracking modalities through plug-and-play open-source interfaces. This flexibility allows developers to adapt their IGT systems depending on the clinical scenario. 

Similarly, robots in any application rely on various algorithms that are often repeated in workflows such as motion planning, inverse kinematics, and simulation. ROS provides users with several libraries and interfaces to access these algorithms without re-implementation. These implementations have also been tested and contributed to by other robotics developers in various domains, which reduces the risk of propagating errors that could potentially damage the robot or introduce instability to the surgical system. 

SlicerROS2 is therefore designed to bridge the tools available in both 3D Slicer and ROS to promote reliable and repeatable construction of these systems. Figure \ref{fig:igtrobot} illustrates this concept.

\subsection{Programming approach}

 Until the initial release of ROS 2 in 2015, ROS was only supported on Linux. ROS 2, which will be the only supported distribution of ROS after 2025, can now be used on Linux, Windows and macOS \cite{macenski2022robot}. This major revision has enabled the opportunity to support more researchers and their applications in ROS, including image-guided robotics. Considering this opportunity, we designed SlicerROS2 to provide access to ROS 2 libraries and features in 3D Slicer. This is done by porting the standard communication mechanisms from ROS 2 into 3D Slicer using rclcpp, which is the C++ API for ROS 2. Although linking libraries as complex as ROS and 3D Slicer is challenging, we chose to implement SlicerROS2 in C++ because it offers better computational performance and better flexibility than the more common integration using Python.

 The following sections outline the specific implementation details and design decisions made for each component of the SlicerROS2 module.

\subsection{ROS overview}
One of the key design elements in ROS (ROS 1 and ROS 2) is the use of multiple processes, called nodes, and various communication mechanisms between these nodes \cite{macenski2022robot}. These communication mechanisms include topics, services, actions, parameters and tf2. A topic acts as a bus for nodes to exchange messages. Topics can be published or subscribed to by nodes. This is done by instantiating a publisher or a subscriber. Separately, parameters are settings for the nodes. Parameters can be used to store integers, floats, booleans, strings and lists. Each node maintains its own parameters \cite{quigley2009ros}. The tf2 library is also an essential feature in ROS 2 for managing 3D transformations between frames in the world such as sensors and robots. For this release, we focused on topics, parameters, and tf2, although the current framework can be extended to support services and actions as needed. 

\subsection{3D Slicer overview} 
In 3D Slicer, Medical Reality Modelling Language (MRML) nodes are used to store all of the data in the scene \cite{fedorov20123d}. This data model was designed to support data types that are required for medical imaging applications. The MRML software library is built on top of the Insight Toolkit (ITK) \cite{mccormick2014itk} and Visualization Toolkit (VTK) \cite{vtkBook}. 

There are several basic MRML nodes that contribute to the core functionality of 3D Slicer. These include MRML data nodes such as: volumes, models, segmentations, markups, transforms, text, and tables. There are also view nodes, display nodes, storage nodes, etc. that are inherent to the basic operation and use of 3D Slicer. Each MRML node has a list of attributes that refer to specific features of that node. As an example, a volume node that contains a CT image will have attributes that include the image spacing and patient orientation. MRML nodes can also have references to other MRML nodes in the scene, i.e., that same volume node can have a reference to a transform node that is used to move the image around the scene. Finally, specific events can be observed on MRML nodes and used to trigger callback functions. 

For custom applications or high level programs, the MRML nodes that are already available in 3D Slicer may not be sufficient. Therefore, custom MRML nodes can be incorporated in loadable 3D Slicer modules. A loadable module is a C++ extension that's compiled against 3D Slicer. 

\begin{table*}[t]
\caption{\emph{Left:} ROS 2 data type. \emph{Middle:} 3D Slicer data type. \emph{Right:} SlicerROS2 class type.}
\begin{center}
\rowcolors{2}{white}{lightgray}
\begin{tabular}{ c c c }
 \textbf{ROS 2} & \textbf{3D Slicer} & \textbf{SlicerROS2 ``name''} \\ 
 \hline
 \codetable{std\_msgs$::$msg$::$String} &  \codetable{std$::$String}  & \codetable{String} \\  
 \codetable{std\_msgs$::$msg$::$Bool} &  \codetable{bool}  & \codetable{Bool} \\
 \codetable{std\_msgs$::$msg$::$Int64} &  \codetable{int}  & \codetable{Int} \\
 \codetable{std\_msgs$::$msg$::$Float64} &  \codetable{double}  & \codetable{Double} \\
 \codetable{std\_msgs$::$msg$::$Int64MultiArray} &  \codetable{vtkIntArray}  & \codetable{IntArray} \\
 \codetable{std\_msgs$::$msg$::$Float64MultiArray} &  \codetable{vtkDoubleArray}  & \codetable{DoubleArray} \\
 \codetable{std\_msgs$::$msg$::$Int64MultiArray} &  \codetable{vtkTable}  & \codetable{IntTable} \\
 \codetable{std\_msgs$::$msg$::$Float64MultiArray} &  \codetable{vtkTable}  & \codetable{DoubleTable} \\
 \codetable{geometry\_msgs$::$msg$::$PoseStamped} &  \codetable{vtkMatrix4x4}  & \codetable{PoseStamped} \\
 \codetable{geometry\_msgs$::$msg$::$WrenchStamped} &  \codetable{vtkDoubleArray}  & \codetable{WrenchStamped} \\
 \codetable{geometry\_msgs$::$msg$::$PoseArray} &  \codetable{vtkTransformCollection}  & \codetable{PoseArray} \\
 \codetable{sensor\_msgs$::$msg$::$Image} &  \codetable{vtkUInt8Array}  & \codetable{UInt8Image} \\
 \codetable{sensor\_msgs$::$msg$::$PointCloud} &  \codetable{vtkPoints}  & \codetable{PointCloud} \\
\end{tabular}

\label{tab:data-types}
\end{center}
\end{table*}

\subsection{Combining ROS and 3D Slicer}

The goal of SlicerROS2 is to provide access to the utility of the MRML scene in ROS 2 and the utility of ROS 2 in 3D Slicer. 3D Slicer is treated as a ROS 2 node and we encapsulate each of the supported ROS communication mechanisms as custom 3D Slicer data types. 

\begin{figure} [h]
    \centering
    \includegraphics[width=9 cm]{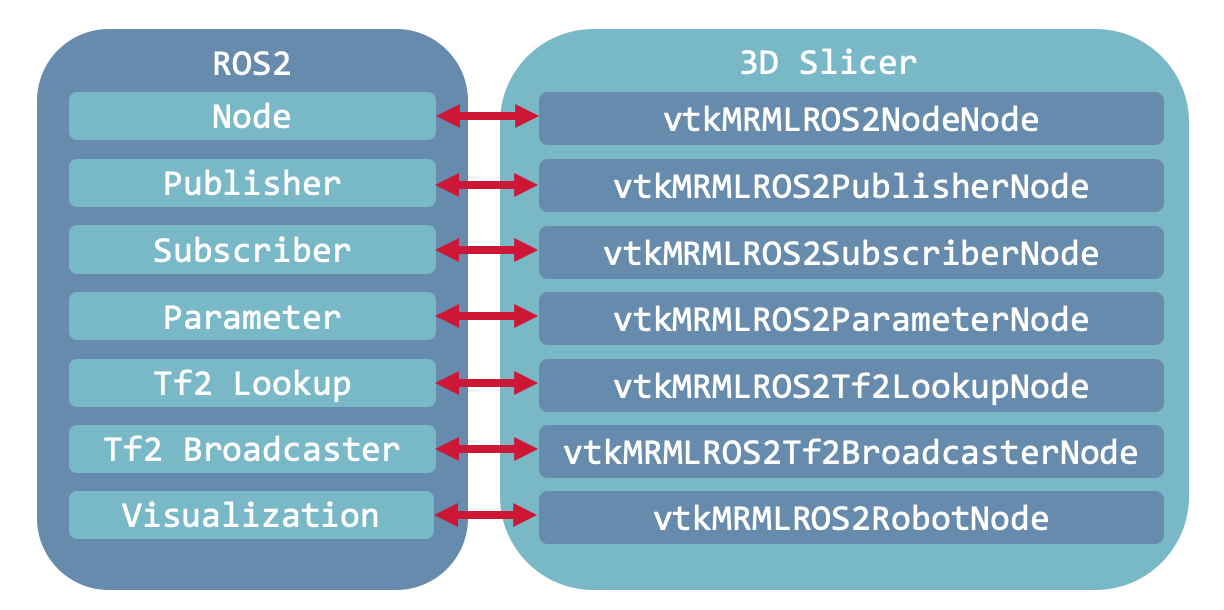}
    \caption{\emph{Left:} ROS 2 elements \emph{Right:} SlicerROS2 / MRML scene equivalent.}
    \label{fig:NodeIntro}
\end{figure}

We also exploit the natural equivalent of the tf2 data structure in 3D Slicer (the transformation hierarchy), for robot visualization. To achieve this, we developed a loadable 3D Slicer module that includes custom MRML nodes for each of the aforementioned ROS 2 elements (Figure \ref{fig:NodeIntro}). 

Each of these new nodes is derived from a \code{vtkMRMLNode} and therefore allows us to leverage some core Slicer features. These features include: using observers to trigger user code when a new ROS message has been received, data visualization, retrieving the node by ID, access to the Python API that is automatically generated on top of native C++ code, and finally, the ability to save and restore ROS 2 nodes along with the MRML scene. Each node follows the 3D Slicer naming convention \code{vtkMRMLxxxxNode} but we add the prefix ROS 2 and use the xxxx to represent the ROS functionality, i.e., \code{vtkMRMLROS2ParameterNode} for a parameter node where xxxx is replaced by \code{ROS2Parameter}. The following sections will describe each of these custom MRML nodes in more detail.

\subsection{Nodes} \label{node}

\begin{figure} [h]
    \centering
    \includegraphics[width=6 cm]{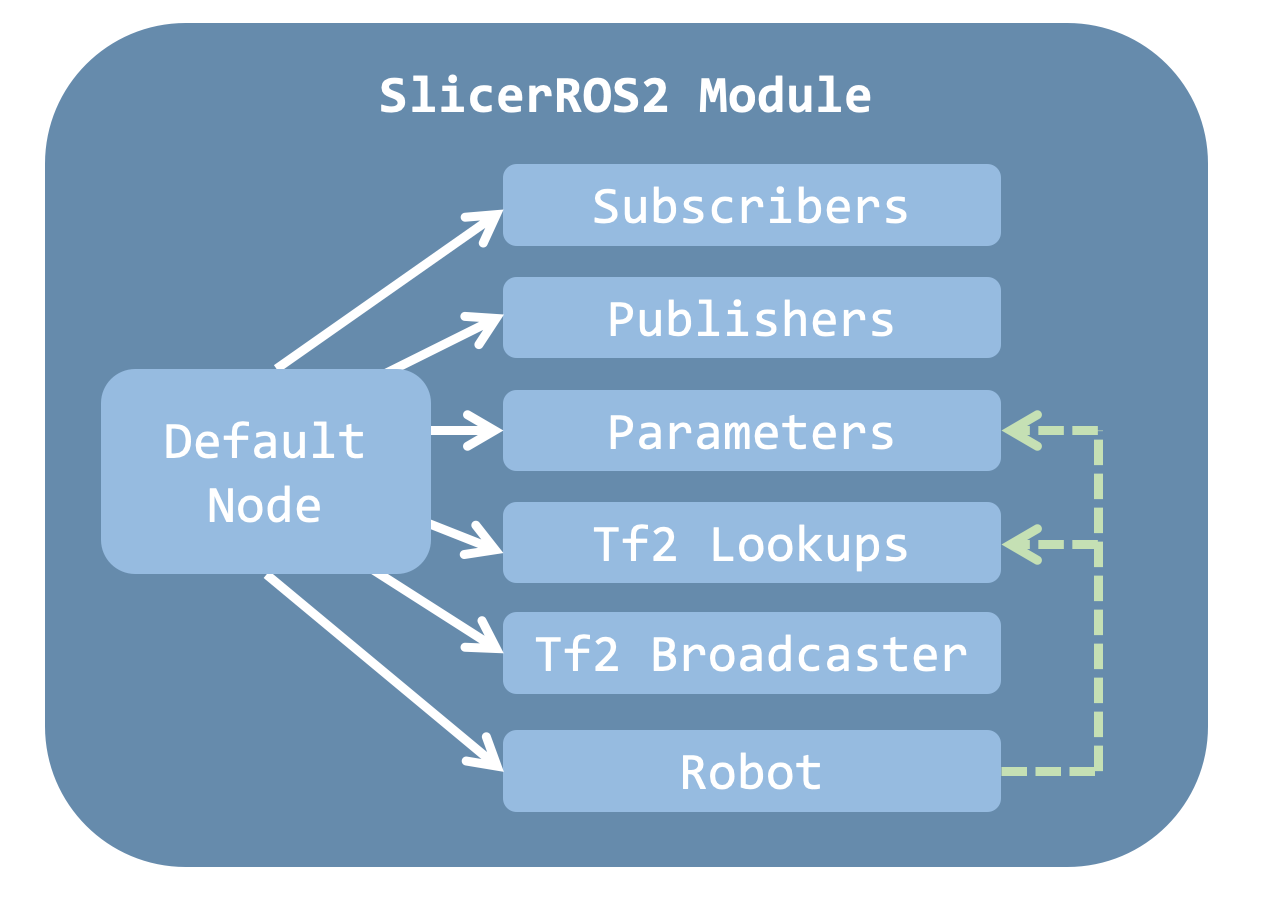}
    \caption{Relationship between the default SlicerROS2 node and other SlicerROS2 nodes.}
    \label{fig:Overview}
\end{figure}

For simplicity, we enforce a default ROS 2 node (\code{vtkMRMLROS2NodeNode}) upon creation of the module to manage the behaviour of publishers, subscribers, parameters, tf2, and robots in SlicerROS2. This relationship is outlined in Figure \ref{fig:Overview}. 

The default SlicerROS2 \emph{node} (from now on we refer to a \code{vtkMRMLROS2NodeNode} as \emph{node} for simplicity) is also used to drive the execution model for the module. As ROS relies heavily on callbacks that are triggered by incoming messages, it is necessary to either use a separate thread to ``spin" the ROS event loop (i.e., check and process new messages), or periodically call a ``spin'' method. Multi-threading in 3D Slicer is not trivial because the MRML scene is not thread-safe. Although some 3D Slicer extensions (i.e., OpenIGTLinkIF) are multi-threaded, it did not seem necessary for this application. Therefore, the SlicerROS2 module relies on existing Slicer mechanisms to trigger a periodic call on the default ROS 2 node to spin. We currently use a Qt timer to trigger this periodic call which is set to 50 Hz or 20 ms by default. This was selected because 50 hz is sufficient for real-time visualization and is also the default refresh rate of tf2. Additionally, any \emph{node} can be used to create, retrieve and remove the other SlicerROS2 MRML nodes (ie. a publisher node can be instantiated by the \emph{node} using the function \code{CreateAndAddPublisherNode}).

\subsection{Data types}

Each communication mechanism in ROS, such as publishers and subscribers, employs a set of predefined data types. These data types are automatically generated from description files that outline the content of the message (\code{.msg}, \code{.srv}, etc). In contrast, 3D Slicer operates on either VTK or its native message types ( i.e., \code{vtkMRMLTransformNode}).

To reconcile these differing data types and ensure smooth interoperability, we introduced several conversion functions. These functions translate messages from ROS 2 data types into their equivalent forms in 3D Slicer, and vice versa. We currently support a variety of standard message types, as illustrated in Table \ref{tab:data-types}. The design of SlicerROS2 also allows developers to extend the software to suit their own use case by providing a simple means for adding new subscribers and publishers. In the C++ code, the user needs to simply define the macro for their new subscriber, write the conversion function in the SlicerToROS2 or ROS2ToSlicer methods and register their new data type in the module logic.

\subsection{Publishers and subscribers}

SlicerROS2 publishers are designed to facilitate data transfer from 3D Slicer to ROS. They can be created and deleted from any SlicerROS2 \emph{node}. To instantiate a new publisher, the user needs to specify both the class name and the topic to which the data will be published. The class name is defined by the syntax: \code{vtkMRMLROS2PublisherXXXXNode}, such that \code{XXXX} is the SlicerROS2 class type (column 3 in Table \ref{tab:data-types}).

SlicerROS2 subscribers are designed to receive data from ROS in 3D Slicer. They have a very similar API to SlicerROS2 publishers.

\subsection{Parameters}

The structure of the SlicerROS2 parameter node differs from other MRML ROS nodes. A typical ROS 2 parameter is distinguished by two components: the hosting ROS node and the unique name of the parameter. This might imply that each individual ROS parameter should have its own individual \emph{node}. However, the ROS 2 libraries facilitate retrieval of all parameters from a single ROS 2 node using just one message. This efficient feature led us to utilize one \emph{node} per parameter hosting ROS 2 node, irrespective of the number of parameters it contains. Therefore, each \emph{node} can monitor all parameters associated with its corresponding ROS 2 node. This design enhances the efficiency of retrieving parameters. 

\subsection{Tf2}

As previously mentioned, tf2 is a functionality in ROS 2 that allows users to track multiple coordinate frames at once. To access transforms in tf2, a listener must be instantiated to parse the transform tree for a particular relationship, given the parent and child ID. The tf2 listener relies on a buffer which tracks all of the transforms that are broadcasted to the ROS 2 graph. A query for a particular transform between frames can then be composed from the buffered data. 3D Slicer manages transforms as \code{vtkMRMLTransformNodes} and offers a visualization of the transformation hierarchy in the built in ``Data" module. To access transformations defined by tf2 in 3D Slicer, a \code{vtkMRMLROS2tf2LookupNode} (which is derived from a \code{vtkMRMLLinearTransformNode}) was developed that queries the tf2 graph for the transformation between two frames. This query / lookup is executed by the spin method on the associated SlicerROS2 \textit{node}. 

A tf2 broadcaster is a similar concept but rather than performing a query, a tf2 broadcaster broadcasts a transformation between two frames to the tf2 graph.

\subsection{Robot}

The Robot Node in SlicerROS2 is designed to facilitate the handling of ROS 2 enabled robots within 3D Slicer. As shown in Figure \ref{fig:Overview}, a SlicerROS2 robot node relies on the tf2 lookup nodes, and a parameter node. The parameter node is used to retrieve the robot description from the robot state publisher, which contains the Unified Robot Description Format (URDF) file. This file can be parsed to get the file path for each STereoLithography (STL) that should be loaded, the offsets of each link from the previous, the colors of the links, frame ids, etc. The robot node is responsible for automatically setting up the robot. This process includes: parsing the URDF file, loading the STL models into the MRML scenes, applying the appropriate offset to each model and finally, setting up and initializing the transformation tree for each link (using SlicerROS2 tf2 lookup nodes). There is also the option to specify the fixed frame that the robot moves with respect to in cases where it may not be the robot's base or the device has non-holonomic constraints.

\begin{figure}[t]
    \centering
    \includegraphics[width=9cm]{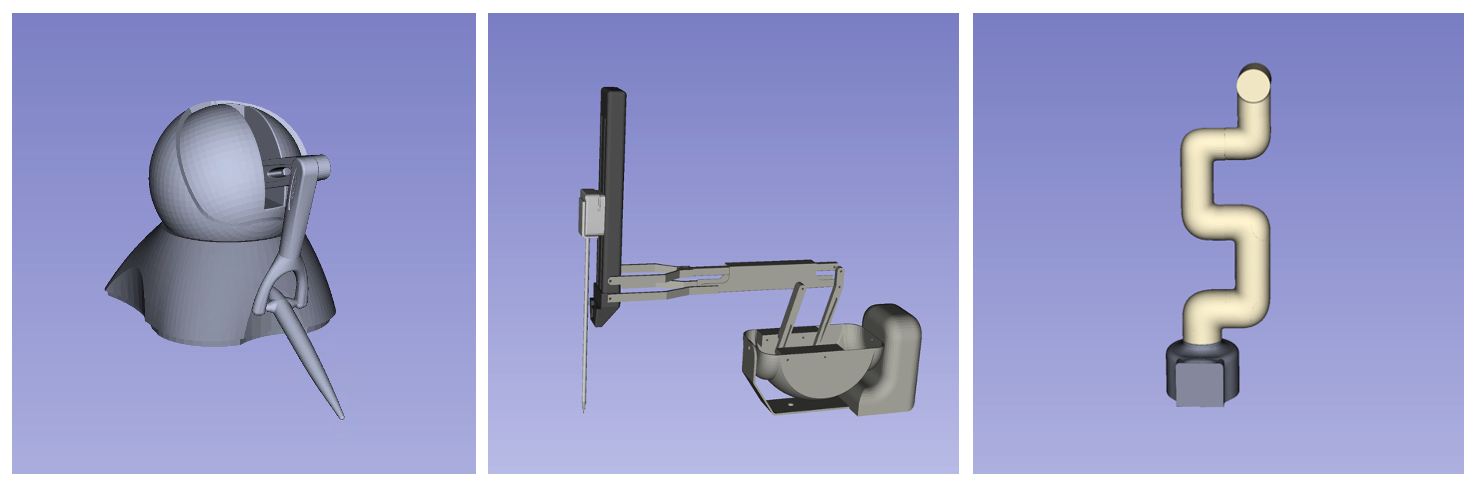}
    \caption{Robots loaded into 3D Slicer via SlicerROS2. \emph{Left:} Omni Bundle robot. \emph{Middle:} da Vinci Research Kit (dVRK) patient side manipulator (PSM). \emph{Right:} myCobot. }
    \label{fig:robos}
\end{figure}

This design enables users to load any ROS 2 enabled robot simply by inputting the name of the parameter node that is receiving the robot description, along with the name of the parameter containing the robot description. Figure \ref{fig:robos} demonstrates three different robots loaded into 3D Slicer using SlicerROS2.

\subsection{API}

As described, the classes provided by the SlicerROS2 module can be accessed through the Python interpreter in Slicer. Below is an example of how to use the API to create a ROS 2 publisher in Slicer, where \code{/pub} is the name of the topic that will be published to: 

\begin{lstlisting}[style=codestyle]
1. logic = slicer.util.getModuleLogic('ROS2')
2. node = logic.GetDefaultROS2Node()
3. publisher = node.CreateAndAddPublisherNode
  ('vtkMRMLROS2PublisherStringNode','/pub')
4. publisher.Publish('Hello world.')
\end{lstlisting}

The workflow is therefore: 1. Retrieve the SlicerROS2 logic, 2. Retrieve the default ROS 2 node, 3. Create a publisher, and finally, 4. Publish to the specified topic. 

\subsection{Performance results}

We first evaluated the performance of SlicerROS2 by computing the time needed to spin the node, process all of the incoming ROS messages and refresh the robot visualization. This was done using a \code{vtkTimerLog} to capture the execution time of each spin step while rendering a robot. Over 100 clock cycles, the default ROS 2 \emph{node} had an execution time of 2.41$\pm$0.79 ms per spin. The frame rate of the 3D viewer in Slicer also did not appear to be impacted by the rendering, maintaining a steady rate of about 49 FPS while the robot was in motion. Rendering an additional robot did not substantially increase this computation time either (3.22$\pm$1.43 ms) and did not change the frame rate of the 3D viewer. These results indicate that we need about 2-3 ms of compute time every 20 ms (the spin rate defined in Section \ref{node}) to avoid interference with other 3D Slicer processes (about 10 - 15 \%\ of the load). 

We also performed the same test on the previous release (the original SlicerROS2 offering \cite{connolly2022bridging}) and discovered that the ``Slicer" node was spinning every 56.48$\pm$1.73 ms while we were requesting a higher refresh rates.  This means that we ran into overruns and the execution time was around 57 ms. This substantial performance improvement (about 20 times faster) is the result of fine tuning the processing steps within the module and only updating the main 3D Slicer thread after each spin cycle, a feature not present in the previous design.

\begin{figure} [h]
    \centering
    \includegraphics[width=9cm]{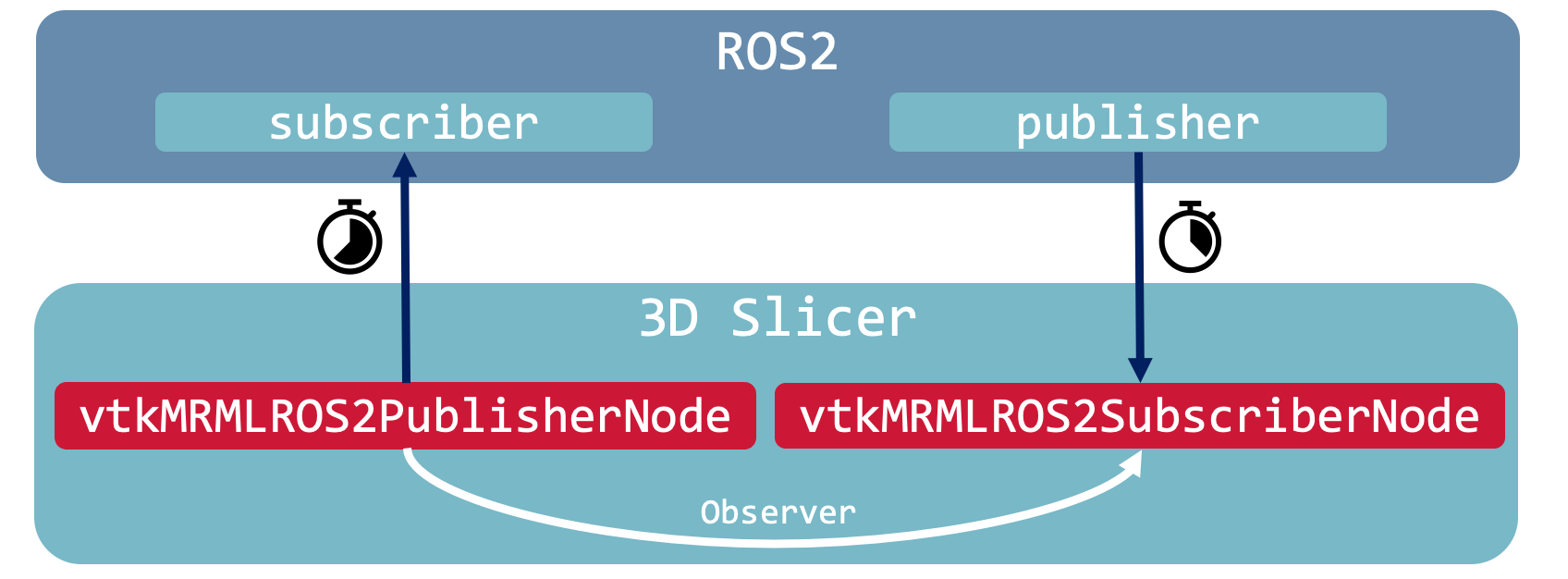}
    \caption{Overview of data transfer test. Messages are sent from ROS 2 to 3D Slicer and back. The built in 3D Slicer observer mechanism is used to relay data after it has been received in Slicer.}
    \label{fig:relaytest}
\end{figure}

In addition to the execution time, we evaluated the data transfer latency of the module by creating two simple python scripts. The first was within SlicerROS2 and acted as a server and the second was  implemented using `rclpy` outside of SlicerROS2, and acted as a client. This test was facilitated in a loop where a message was sent from the ROS 2 client to the 3D Slicer server and then back to ROS 2. To do this, we instantiated a floating point publisher and subscriber in ROS 2 as well as a SlicerROS2 subscriber and publisher. The SlicerROS2 publisher was programmed to send a message back to the client each time the SlicerROS2 subscriber node received new data using the MRML node observer and callback infrastructure. For the test, we published the current time with the ROS 2 publisher, subscribed to this topic with the SlicerROS2 subscriber, which then triggered the SlicerROS2 publisher to relay the data back to ROS 2. The data was then captured by the ROS 2 subscriber and we computed the total ``time of flight" by comparing the current time when the message was received to the time when the message was sent. Figure \ref{fig:relaytest} illustrates the loop that was used to execute this test. 

We determined that when both the ROS 2 node and the SlicerROS2 \emph{node} were set to 50 hz (20 ms) the average time of flight was 11.9$\pm$0.3 ms over 100 messages. Moreover, when the ROS 2 node and SlicerROS2 \emph{node} were set to 100 hz (10 ms) the average time of flight was 8.5$\pm$0.3 ms. We also noted that all of the messages that were sent were also received in both tests. As anticipated, the time of flight is about half of the data transmission rate. Currently the speed limiting factor for SlicerROS2 is the refresh rate of the Slicer 3D renderer and built-in observer mechanisms on the MRML nodes. These tests were performed on a 10 core hyper threaded i7-12700k processor with a 3.6 GHz clock, 64 GB of RAM, an RTX 3090 and no other processes running. The system was set up with Ubuntu 20.04, ROS2 Galactic and a release build of Slicer 5.1.0. \\

\section{Example applications \&\ use-cases}

\begin{figure*} [h]
    \centering
    \includegraphics[width=18cm]{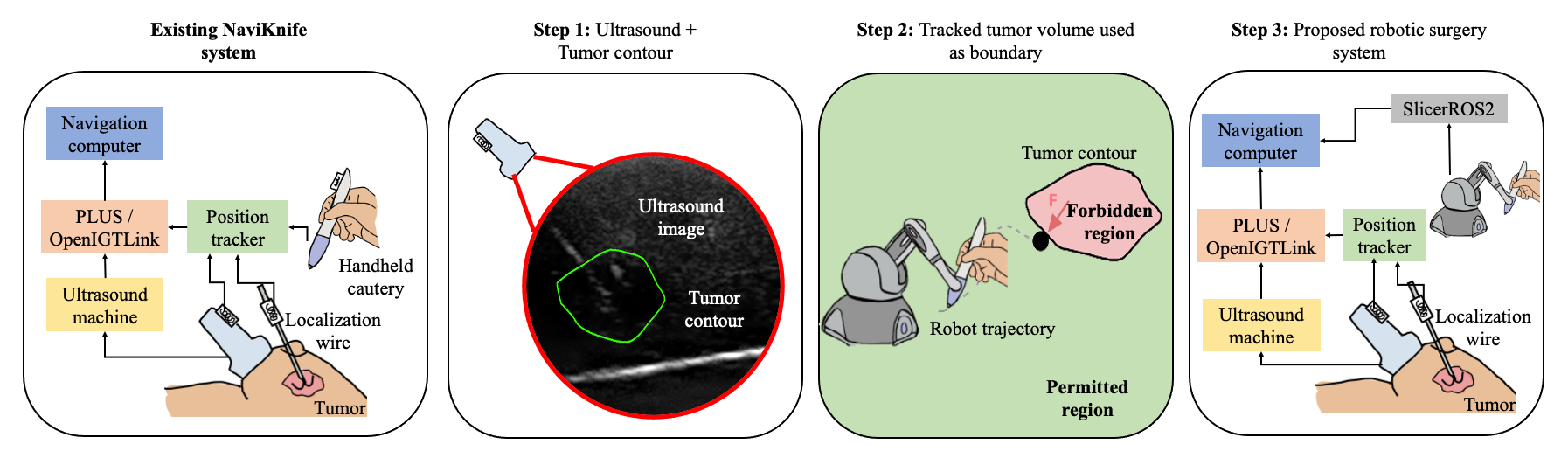}
    \caption{Overview of existing NaviKnife system and proposed steps to include VF guidance. \emph{Left:} Existing NaviKnife system including equipment and information flow. \emph{Step 1:} Identification of tumor contour from ultrasound. \emph{Step 2:} Visualization of forbidden region that is imposed around tumor contour. \emph{Step 3:} Full cooperative robotic system with integrated VF.}
    \label{fig:naviKnife}
\end{figure*}

To demonstrate the utility of SlicerROS2, the following sections will describe two potential, clinically relevant use-cases of the platform. Specific quantitative results of these use-cases would be reflective of a combination of factors including registration accuracy or hardware quality and not indicative of the performance quality of the software described in this paper. Therefore, we describe high-level results that are indicative of the module performance and data pipeline.

\subsection{Virtual fixtures}
A virtual fixture (VF) is a constraint communicated from robots to operators via force and position signals \cite{abbott2007haptic}. In medical robotics, VFs are often used to prevent the user (surgeon) from entering a forbidden region, like a critical piece of anatomy, or to guide them along a desired path \cite{moccia2020vision} \cite{duan2021virtual}. In 3D Slicer, registration and localization of such critical anatomy can be done with some of the built-in tools and extensions like ``Fiducial Registration Wizard'' or the ``Segmentation Editor''. Moreover, the ``Breach Warning'' module, which is packaged as part of the SlicerIGT extension, can be used to compute the distance of a particular transform in the scene to the surface of this critical anatomy. This module instantiates a breach warning node which can be queried to detect if a certain transform has breached an object. This module was initially developed for a navigation platform called NaviKnife that was designed for breast conserving surgery (BCS) \cite{ungi2015navigated}. BCS is a procedure where a surgeon removes a tumor from the breast with the goal of preventing positive margins, or tumor breaches. The NaviKnife platform relies on preoperative ultrasound to identify tumor boundaries, and electromagnetic tracking to track the tumor via a localization needle that is inserted into the tumor before surgery. The surgeon's electrocautery is then tracked relative to the tumor and a sound or color change is used to indicate when the cautery has breached the tumor. Currently, the NaviKnife platform is deployed in 3D Slicer using OpenIGTLink and the PLUS toolkit \cite{lasso2014plus} to stream the ultrasound and EM tracking data into Slicer. Adding a VF to this platform as an additional method for preventing tumor breach can be done by setting the tumor boundary being identified in ultrasound as the boundary for applying haptic feedback. Additionally, the surgeon's electrocautery can be cooperatively guided by a haptic device rather than hand-held (Figure \ref{fig:naviKnife}).

Given a model of the tumor and the tracking information in 3D Slicer (as provided by NaviKnife), adding VF guidance to this platform requires an additional connection to the Omni Bundle robot through SlicerROS2. The VF functionality can then be achieved with only a few lines of Python code (Algorithm 1).

\begin{algorithm}
\caption{Virtual fixture logic / pseudo code}\label{euclid}
\begin{algorithmic}[1]
\State Retrieve the SlicerROS2 module logic 
\State Retrieve the default SlicerROS2 node from the logic
\State Create a SlicerROS2 subscriber to the robot pose (S1)
\State Create a SlicerROS2 publisher to the robot pose (P1)
\State Create a SlicerROS2 publisher to the robot wrench (P2)

\State Retrieve the breach warning node (BWN) from the scene (to determine if the tool has breached the tumor)
\State Add an observer to the BWN to check when it is modified and trigger the following conditional statement:

\If{tool has breached the tumor}
    \State Get the cartesian position of the robot from S1
    \State Publish the cartesian position of the robot to P1
\Else{}
    \State Publish a null wrench to P2
\EndIf
\end{algorithmic}
\end{algorithm}

\begin{figure} [h]
    \centering
    \includegraphics[width=8.5cm]{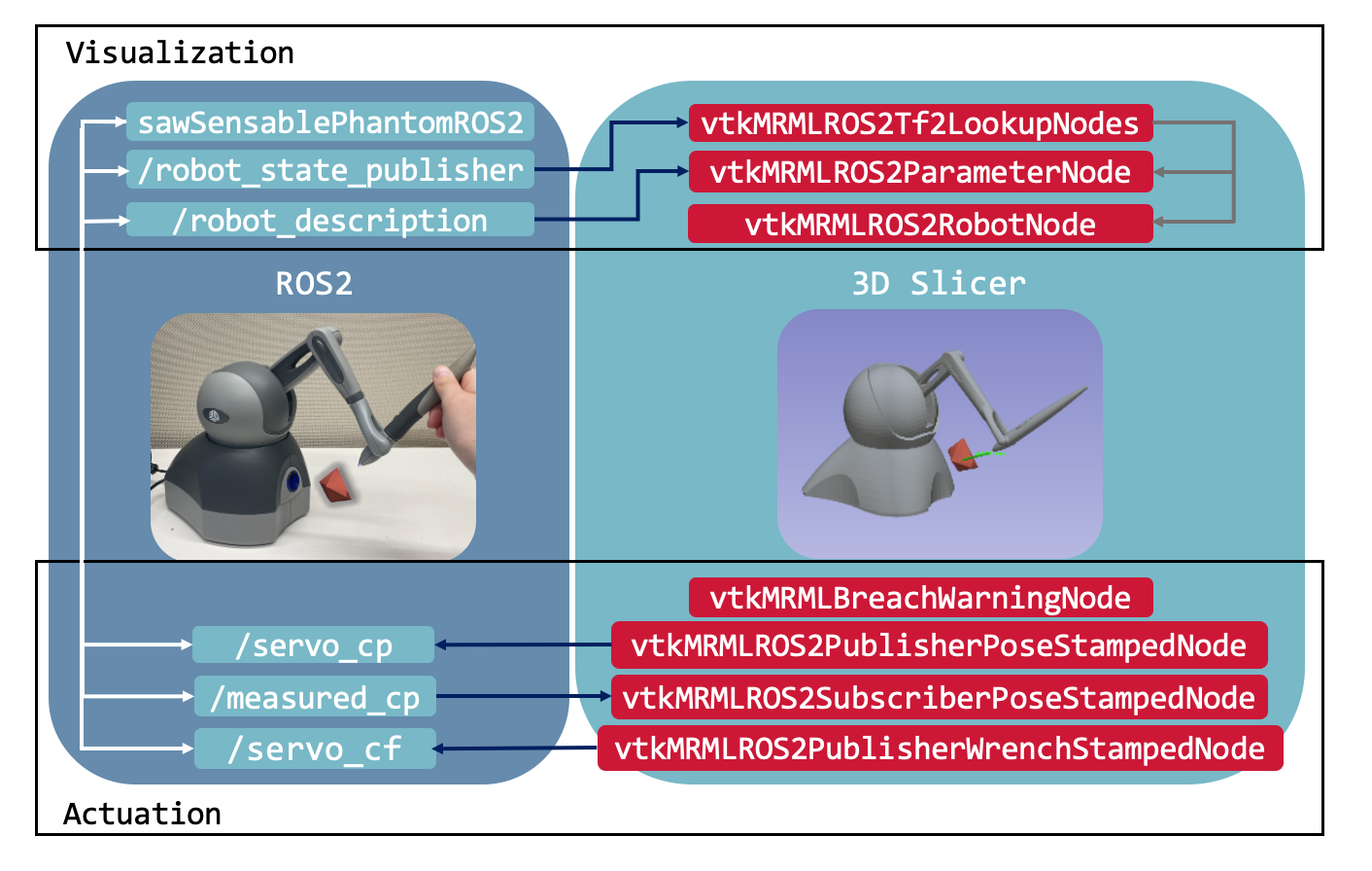}
    \caption{Overview of data transfer between ROS 2 and 3D Slicer to deploy VF example. On the left, topics and node names are shown alongside an image of the Omni with the a visual of the virtual tumor. On the right, custom SlicerROS2 nodes are shown alongside a rendering of the robot in the 3D Slicer viewer.}
    \label{fig:VF}
\end{figure}

 With this algorithm in place, the user will feel a ``hold in place'' or ``sticky'' haptic feedback when the tip of their cooperatively held tool breaches the tumor boundary. This logic was tested using the Omni Bundle robot (formerly known as the Phantom Omni - Quanser, Markham, ON, Canada) and a pseudotumor shown in Figure \ref{fig:VF}. The ROS 2 interface for the Omni bundle is provided by sawSensablePhantomROS2 \cite{sawSensable} package which is part of the SAW libraries. More details on this communication are available in \cite{connolly2022bridging}. All of the ROS 2 and SlicerROS2 counterparts necessary to set up the VF are shown in Figure \ref{fig:VF}. Streaming of the ultrasound and EM tracker data is still facilitated with PLUS and OpenIGTLink because the devices used in the NaviKnife system do not have open-source ROS wrappers. The Omni Bundle robot is integrated seperately with SlicerROS2. To test this implementation, we asked a small group of users (n=4) to move freely through the robot's workspace for 30 seconds and indicate when they felt haptic feedback without looking at the visualization. When they felt resistance, we asked them to hold the robot in place and placed a fiducial at the tip of the end-effector in order to capture the position where the feedback was felt. Only one user placed 2 fiducials (out of 5 total) outside of the tumor region. These fiducials were placed 2.0 and 3.4 mm from the surface of the tumor and we believe that was due to inadvertent motion between the time of indication and the placement of the fiducial. These results therefore indicate that SlicerROS2 can be used to define virtual fixtures and enforce them, as intended. This specific application was also presented as part of the Hamlyn Symposium on Medical Robotics (HSMR 2023) \cite{connolly2023slicerros2}.
 
This example shows that incorporating haptics into image-guided procedures can now be prototyped without a custom interface and with minimal developer intervention. This approach is much simpler than previous implementations of ultrasound-based virtual fixtures developed with custom C++ software \cite{alamilla2022}. This application is also significant because the Omni Bundle robot is a relatively inexpensive device that is often used as a teaching tool. Considering this, researchers can use this example as a starting point to begin researching and applying cooperative robotics with minimal startup cost and development hours.

\subsection{Image guided motion planning}
\begin{figure*} [t]
    \centering
    \includegraphics[width=17cm]{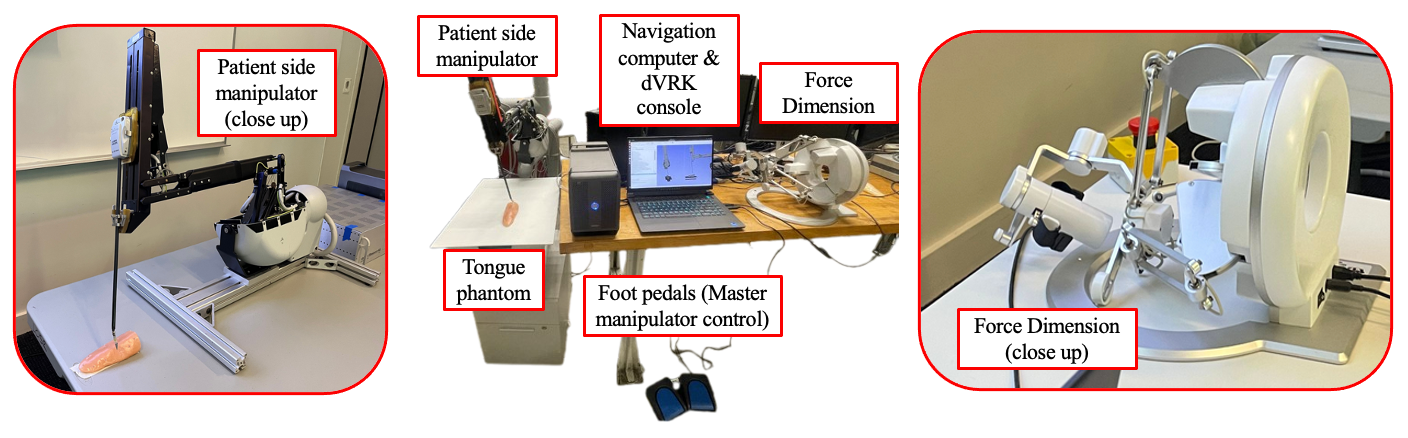}
    \caption{Overview of the hardware setup for the simulated TORS procedure. PSM is shown on the left with the tongue phantom. The navigation computer hosting SlicerROS2 and the dVRK console is shown in the center with the foot pedals below. And finally, the Force Dimension is to the right of the navigation computer.}
    \label{fig:hardwareSetup}
\end{figure*}

Image-guidance can also be used to inform path planning and intraoperative visualization for medical robotic systems. To demonstrate how SlicerROS2 can be used in this capacity, we use tongue tumor resection as an example application. Please note that this application was also demonstrated at the International Symposium for Medical Robotics (ISMR 2023) in Atlanta, Georgia \cite{ismr} (Figure \ref{fig:hardwareSetup}). 

\begin{figure} [h]
    \centering
    \includegraphics[width=6cm]{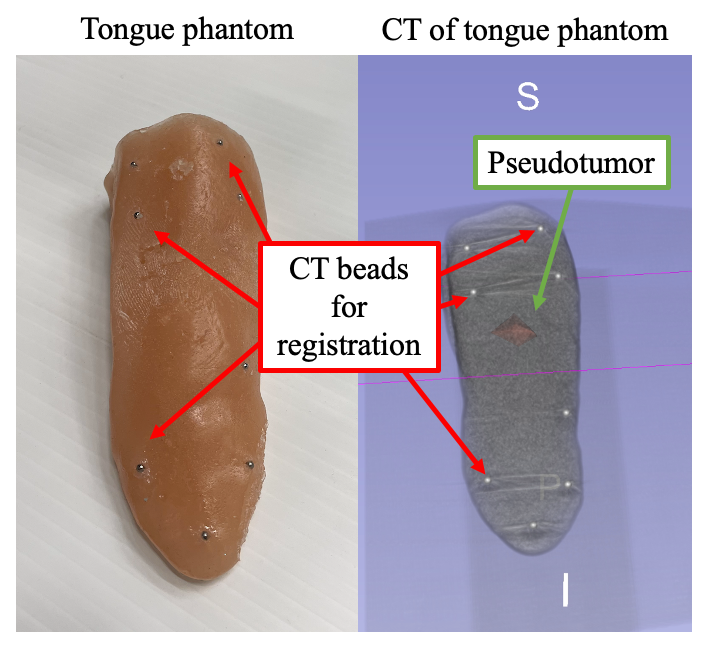}
    \caption{\emph{Left:} Tongue phantom with CT beads on the surface. \emph{Right:} Volume rendering of the tongue in 3D Slicer based on CT images. Pseudotumor shown inside of the tongue.}
    \label{fig:tongue}
\end{figure}

A preoperative computed tomography (CT) image is often used for tongue tumor resection procedures to visualize the tongue and the margins of the tumor \cite{yoon2020comparison}. Alongside imaging, transoral robotic surgery (TORS) is used to treat oral cancers because it is minimally invasive and offers several advantages over open-surgery \cite{o2006transoral}. Using a system like the da Vinci robot has benefited TORS procedures because of improved visualization and access to the oral cavity \cite{van2020one}. To simulate a TORS scenario, we created a tongue phantom out of ballistic gel and placed CT beads along the tongue for registration. The tongue was then imaged using the Brain Lab Loop-X (Munich, Germany) device and rendered in 3D using the ``Volume Rendering'' module in 3D Slicer (Figure \ref{fig:tongue}). We also placed a pseudotumor in the phantom as a target for resection.

We use the da Vinci Research Kit (dVRK) patient side manipulator (PSM) and a Force Dimension Sigma 7 haptic robot (Force Dimension, Nyon, Switzerland) as the master manipulator \cite{kazanzides-chen-etal-icra-2014}. External foot pedals are used as the clutch for the device. SlicerROS2 and the dVRK console are hosted on a laptop between the Force Dimension and PSM as shown in Figure \ref{fig:hardwareSetup}. The end-effector of the dVRK was modified to hold a surgical blade so it could be used to cut through the tongue phantom.

\begin{figure} [h]
    \centering
    \includegraphics[width=8.5cm]{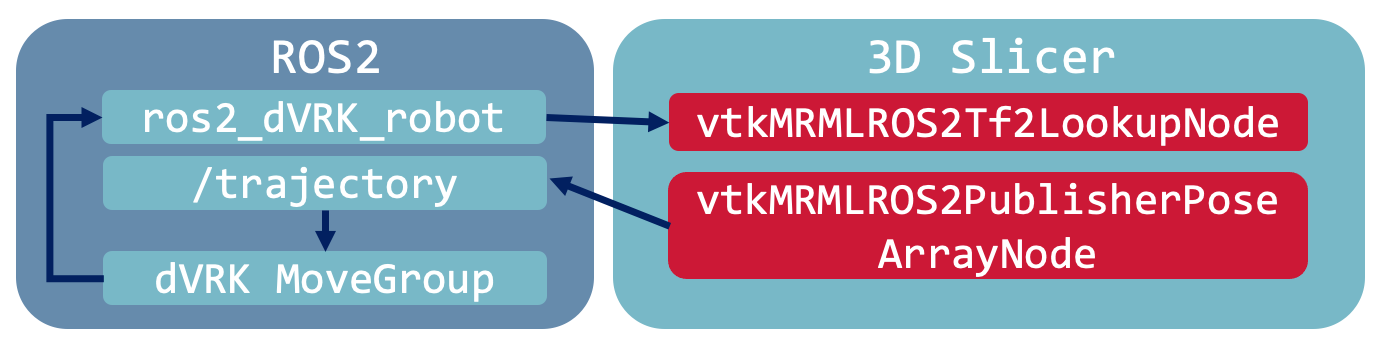}
    \caption{Data transfer between 3D Slicer and ROS 2 to enable trajectory tracing (defined in Slicer) in MoveIt 2. Please note that the visualization of the dVRK is done in the same way as shown in Figure \ref{fig:VF} so the topics and MRML nodes used for visualization are omitted in this example.}
    \label{fig:dataTransfer}
\end{figure}

For this demonstration, we rely on the SlicerROS2 publisher node. Specifically, a \code{vtkMRMLROS2PublisherPoseArrayNode} is used to convert a \code{vtkTransformCollection} to a ROS2 \code{PoseArray}. This \code{PoseArray} is then converted to a trajectory using the dVRK MoveGroup via MoveIt 2. This data transfer protocol is shown in more detail in Figure \ref{fig:dataTransfer}.

\begin{figure} [h]
    \centering
    \includegraphics[width=8cm]{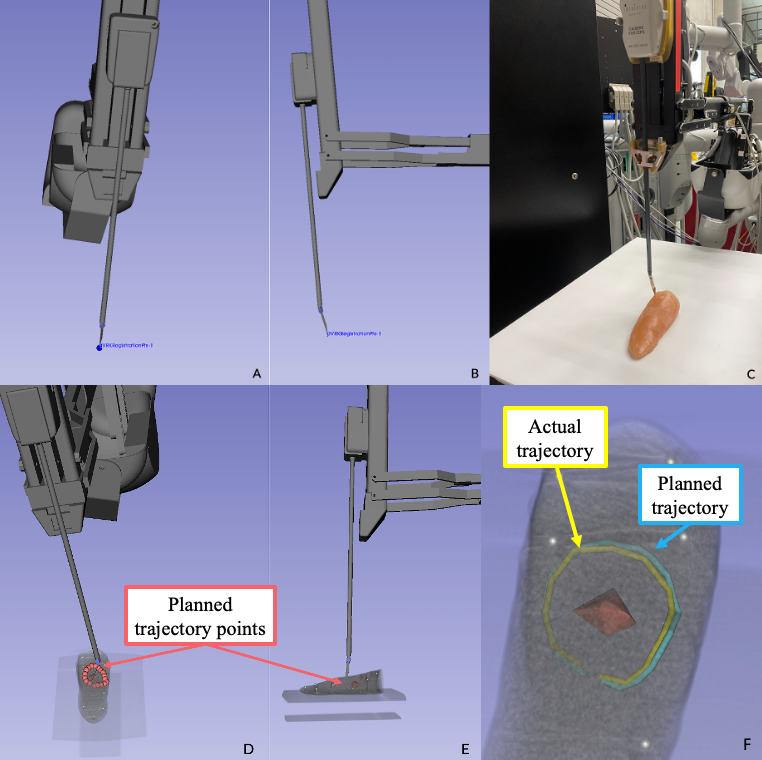}
    \caption{Experiment overview. \emph{A, B:} Front view and side view of the PSM visualization with SlicerROS2. A blue fiducial is shown on the tip of the robot which was used as the registration point for the CT bead in the dVRK coordinate system. This fiducial was placed when the tip of the dVRK was on each CT bead and its position is based on the internal kinematics of the robot. \emph{C:} Real world view of the dVRK on top of a physical CT bead during registration. \emph{D, E:} Front and side view of the dVRK and its planned trajectory traced with the simulated arm. \emph{F:} Front view of the planned (shown in blue) versus the actual trajectory (shown in yellow) of the PSM tip.}
    \label{fig:IGTReg}
\end{figure}

For this experiment, we first registered the PSM to the tongue phantom by selecting the points where the CT beads are in the volume rendering (image coordinate system) and guiding the tip of the PSM to the same points on the tongue phantom (PSM coordinate system), as shown in Figure \ref{fig:IGTReg} - A, B and C. The ``Fiducial Registration Wizard'' module was used to generate the Transform ``CTtodVRK'' which was then applied to the volume rendering and pseudotumor so that the tongue rendering could be visualized in the robot's coordinate system (Figure \ref{fig:IGTReg}). We then switched the dVRK console to control a simulated PSM with the Force Dimension instead of the real PSM, and asked a manipulator to move this simulated PSM in a circular path around the pseudotumor using the built in 3D Slicer viewers (Figure \ref{fig:IGTReg} - D, E). While the user traced the path, we tracked the position of the simulated PSM to create a desired trajectory trace and placed digital markups along the trajectory so they could see the cut path in the 3D Slicer view. We also captured a \code{vtkTransformCollection} that was comprised of the pose of the end-effector of the PSM at each point along the path. The pose array publisher and data transfer protocol described in Figure \ref{fig:dataTransfer} was then used to send the dVRK along this trajectory. 

Figure \ref{fig:IGTReg} - F shows the planned trajectory compared to the actual trajectory followed by the PSM (as defined by the measured pose) around the pseudotumor. With this navigation view, the manipulator could successfully trace a path around the pseudotumor which can be closely followed by the actual robot. The Euclidean distance between the two trajectories was $2.75\pm0.65$ mm.

This example outlines the simplified workflow that is now available with SlicerROS2 to use medical image data to inform motion planning and execution. Although we asked an operator to trace the path around the pseudotumor, deep learning or image processing methods could be implemented to identify this path autonomously. Therefore this module has the potential to accelerate research prototyping of autonomous robotic systems that rely on imaging for context. Previously, the same workflow would require bridging to achieve data fusion between 3D Slicer to ROS. For example, a registration done in 3D Slicer would need to be exported to ROS, or a planned trajectory computed in ROS would need to be imported into 3D Slicer. This process would likely be facilitated with middleware, which can increase latency and complexity. Fusing all of this data within the SlicerROS2 module can, therefore, reduce the cost, time and effort associated with prototyping image-guided robotic systems.

\subsection{Robot and image simulation}
Dynamic simulation is an essential tool for prototyping and testing robotic systems. One of the most popular systems for dynamic robot simulation is Gazebo (formerly known as Ignition), which has several features such as dynamic loading, advanced graphics, realistic physics computation, plugin prototyping and ROS integration \cite{koenig-howard-iros-2004}. Gazebo currently supports the simulation of several sensors, such as LIDAR, GPS and IMU, which are commonly used for mobile and industrial robotics, but does not support sensors commonly used for image-guided robotics, such as ultrasound, MRI or CT. Although Gazebo has a separate visualization GUI, it does not have specialized tools for medical imaging tasks like image segmentation or registration. Therefore, we illustrate an example that uses Gazebo to simulate robotic ultrasound and SlicerROS2 for visualization and planning.

\begin{figure} [h]
    \centering
    \includegraphics[width=8.5cm]{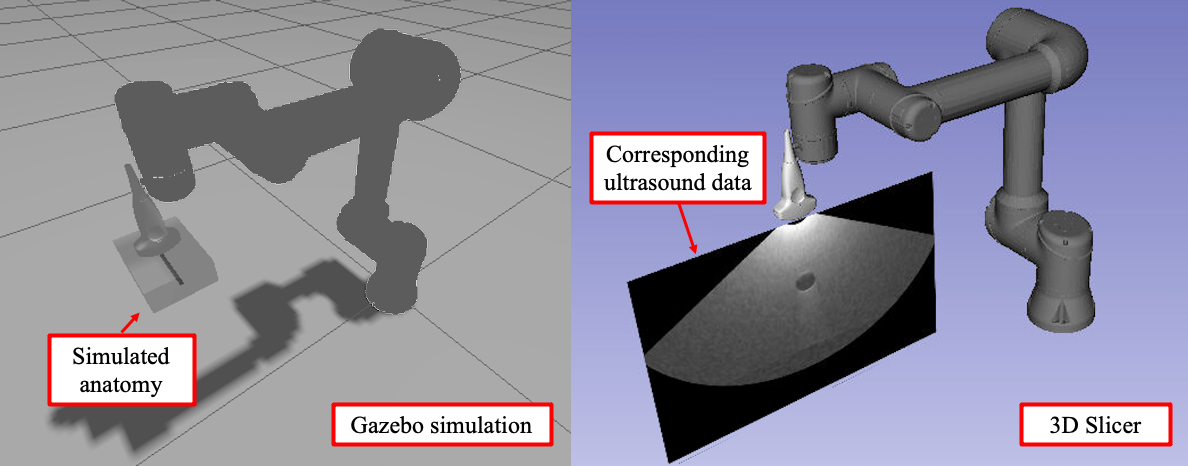}
    \caption{\emph{Left:} Gazebo simulation of robot with simulated anatomy, a vessel in a gel block. \emph{Right:} Corresponding visualization in Slicer, with live simulated ultrasound data.}
    \label{fig:simulatedUS}
\end{figure}

To add the medical sensing capabilities, a plugin to generate  simulated ultrasound images was developed using a surface mesh-based simulation approach demonstrated in \cite{bartha2013open}.  Using this plugin and the ROS 2 interface in Gazebo, the simulated ultrasound data was then published to ROS 2  \code{sensor\_msgs$::$msg$::$Image} and a \code{vtkMRMLROS2SubscriberUInt8ImageNode} was used to receive the data in 3D Slicer. The ``Volume Reslice Driver" module in SlicerIGT was used to show the simulated ultrasound data at the end-effector of the robot in the 3D viewer (Figure \ref{fig:simulatedUS}).

With the synchronized simulation data and 3D Slicer environment, the user can plan a robotic ultrasound intervention with realistic tissue interactions and imaging inputs. We also want to highlight that other simulation engines such as \cite{fontanelli-selvaggio-etal-biorob-2018} \cite{xu-li-etal-iros-2021} and \cite{munawar-wu-etal-ral-2022} can be similarly integrated with SlicerROS2 if they are compatible with ROS 2.
\subsection{Custom device integration and support}

\begin{figure} [h]
    \centering
    \includegraphics[width=8.5cm]{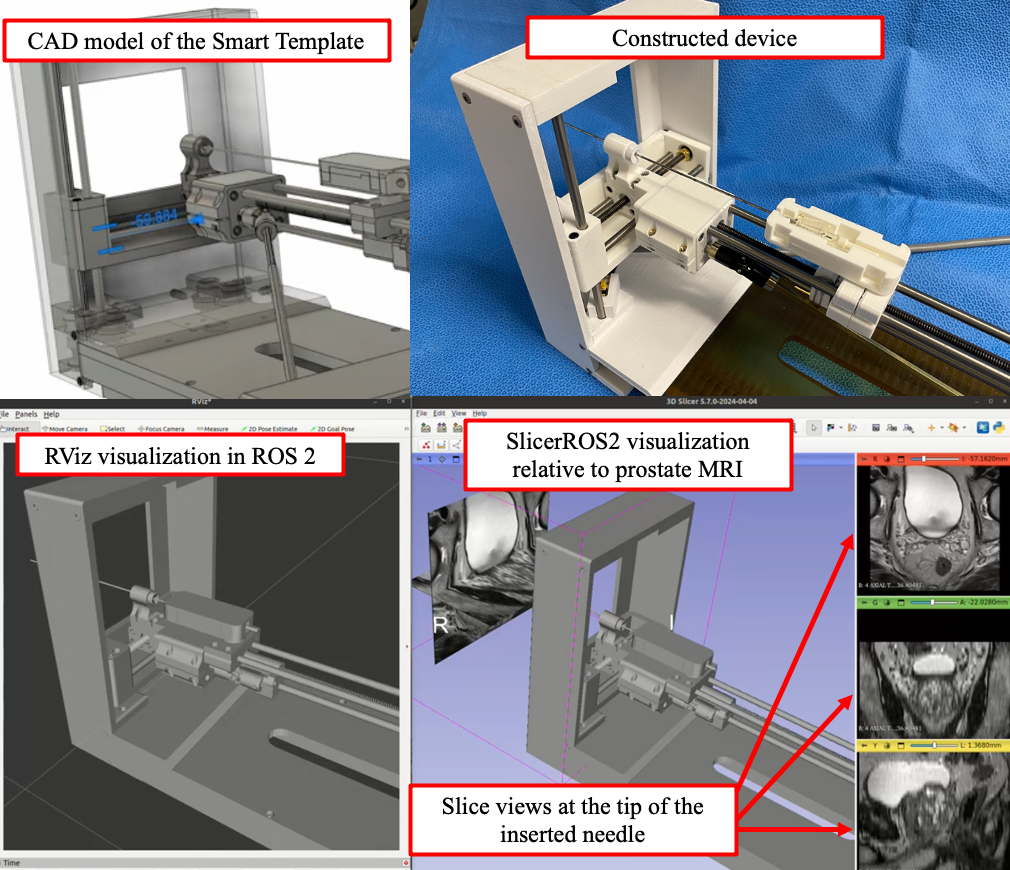}
    \caption{Integration of Smart Template Device with 3D Slicer using SlicerROS2. \emph{Top left:} Visualization of the device in CAD. \emph{Top right:} Image of the manufactured device. \emph{Bottom left:} Visualization in RViz viewer. \emph{Bottom right:} Smart Template rendered with SlicerROS2 relative to MRI prostate images. Slice views included to show cross section of the needle in various imaging planes.}
    \label{fig:SmartTemplate}
\end{figure}

Although many robots used for medical research come with off-the-shelf ROS interfaces, we also recognize that several active research areas rely on custom device development. The design and development of custom devices are particularly prevalent in MRI-guided systems because they must be made from MRI-compatible material, fit within the confines of an MRI scanner, and use non-magnetic motors and actuators, unlike many industrial robots \cite{li2020body}. To illustrate how such a device can be integrated with SlicerROS2, we use a needle-guidance robot called the Smart Template as an example (Figure \ref{fig:SmartTemplate}). The Smart Template is designed for transperineal prostate biopsy to mitigate needle deviation during insertion \cite{moreira2021vivo}. As this system is used for MRI-guided biopsy, visualizing the robot relative to the anatomy at the needle tip is highly useful for guidance and surgical decision-making.

To use this robot with SlicerROS2, a URDF file was generated based on the CAD description of the system. This file describes the structure of the robot (links and joints) and can be generated by measuring and weighing the robot or using a URDF plugin in software like SolidWorks (\url{https://www.solidworks.com/}) or Fusion 360 (\url{https://www.autodesk.com/products/fusion-360}).  This description was then published to ROS 2 to load the robot in 3D Slicer and a joint state publisher was used for moving the device (Figure \ref{fig:SmartTemplate}). An MRI image was also registered to the device, and ``Volume Reslice Driver'' was used to align the image with the needle tip transformation to highlight how visualization in SlicerROS2 can be used for intraoperative planning.

This example demonstrates how, with SlicerROS2, developers do not need to build models for fabrication, control, treatment planning and visualization separately. This feature is useful when a user is prototyping a custom device like the Smart Template and improving it iteratively through design, fabrication, and testing cycles. Without SlicerROS2, a designer would need to individually update the CAD model, kinematics model (both in ROS for control and 3D Slicer for treatment planning), and visualization model at each design iteration. After a few iterations, this manual process could lead to design inconsistencies which SlicerROS2 would help mitigate.

\section{Discussion}

Developing image-guided robotic systems relies on several processes such as visualization, planning, and real-time navigation. Researchers have benefited greatly from well-established tools that provide these features like 3D Slicer and ROS. In this paper, we describe an extensive rework of SlicerROS2 which is an extension of our initial offering in \cite{connolly2022bridging}. We show how this system can be used for applications such as virtual fixtures, image-guided motion planning, simulation and custom device support. Other potential use cases of the platform include integration of robots and augmented / virtual reality support in 3D Slicer \cite{pose2023real}, a fusion of robotics, image-guidance and biomechanical tissue modelling systems \cite{tymkovych2021application} and data-driven medical robotics research with tools like MONAI and 3D Slicer \cite{diaz2022monai}.

\begin{table*}[t]
    \centering
    \caption{Feature comparison between SlicerROS2, the previous release, the ROS-IGTL bridge and custom robot IGT applications. }
    \begin{tabular}{|p{4.5cm}|c|c|c|c|}
        \hline
        \textbf{Features} & \textbf{SlicerROS2}  & \textbf{Previous release \cite{connolly2022bridging}} & \textbf{ROS-IGTL Bridge \cite{tauscher2015openigtlink}} & \textbf{Custom applications \cite{connolly2021open} \cite{frank2017ros}} \\ \hline
        Bi-directional data transfer & $ \cellcolor{mygreen}\checkmark$ & $\cellcolor{mygreen}\checkmark$ & \cellcolor{mygreen}$\checkmark$ & \cellcolor{mygray}$\times$ \\ \hline
        Automatic loading and visualization & \cellcolor{mygreen}$\checkmark$ & \cellcolor{mygreen}$\checkmark$  & \cellcolor{mygray}$\times$ & \cellcolor{mygray}$\times$ \\ \hline
        More than 5 supported data types & \cellcolor{mygreen}$\checkmark$ & \cellcolor{mygray}$\times$  & \cellcolor{mygray}$\times$ & \cellcolor{mygray}$\times$ \\ \hline
        Handling of multiple robots at once & \cellcolor{mygreen}$\checkmark$ & \cellcolor{mygray}$\times$  & \cellcolor{mygray}$\times$ & \cellcolor{mygray}$\times$ \\ \hline
        Python interaction and prototyping & \cellcolor{mygreen}$\checkmark$ & \cellcolor{mygray}$\times$  & \cellcolor{mygray}$\times$   & \cellcolor{mygray} $\times$ \\ \hline
        Access to low level features & \cellcolor{mygreen}$\checkmark$ & \cellcolor{mygray}$\times$  & \cellcolor{mygray}$\times$   & \cellcolor{mygray} $\times$ \\ \hline
        Consistent API & \cellcolor{mygreen}$\checkmark$ & \cellcolor{mygray}$\times$  & \cellcolor{mygray}$\times$   & \cellcolor{mygray} $\times$ \\ \hline
        Active maintenance and documentation & \cellcolor{mygreen}$\checkmark$ &\cellcolor{mygray}$\times$  & \cellcolor{mygray}$\times$   &  \cellcolor{mygray}$\times$ \\ \hline
    \end{tabular} 
    \label{tab:featurecomparison}
\end{table*}

Several changes in this release of SlicerROS2 improve upon previous releases and related works for bridging robot and medical imaging information. These changes are highlighted in Table \ref{tab:featurecomparison}.

Although the previous release of SlicerROS2 and the ROS-ITGL bridge also support bi-directional data transfer, we note that there are only 5 data types supported in the ROS-IGTL bridge (string, transform, image, poly data, and point) and 1 supported in the previous release (a tf2 transform)  whereas SlicerROS2 currently supports 13 different data types. The custom MRML nodes available in SlicerROS2 also greatly simplify this data transfer process.

Based on these new features and improvements, there are several development advantages to using SlicerROS2 over alternative approaches like the ROS-IGTL bridge and custom applications. However, we note that one of the limitations of this module is the reliance on a ROS interface for the device. There are many use cases where a sensor or input, such as an EM tracker or ultrasound, may already have an OpenIGTLink interface, and alternative approaches could be faster and simpler to integrate. In these cases, we are not suggesting that these devices should be ported to ROS for integration but rather contributing a new toolset for devices that already have an open-source ROS interface or could benefit from the addition of one.

There are still ROS features that have not been implemented in SlicerROS2, such as services and actions. Integration of these features and extension of the existing platform will be largely driven by community need. With this, additional feature development will likely target specific use-cases that are built with SlicerROS2. Given this development model, an important aspect of demonstrating the utlity of SlicerROS2 relies on building a community of users and disseminating the platform at conferences and workshops. As previously mentioned, in 2023, a workshop for SlicerROS2 was organized at the International Symposium on Medical Robotics in Atlanta, Georgia with around 30 participants from various universities and research centers. We will continue to provide workshops and presentations to encourage other centers to user SlicerROS2 for their research.

The source code for SlicerROS2 is publicly accessible on Github, licensed under the open-source MIT license. It can be accessed at \cite{slicer_ros2_module}. Additionally, supporting documentation for this project is available on ReadTheDocs, which can be accessed at \cite{slicer_ros2_docs} and a Docker image that was created for the ISMR workshop at \cite{ismr}. Using a similar model as that adopted by PLUS toolkit and OpenIGTLink, the core developers of the module will work to ensure that the code remains up-to-date and address new feature requests and issues.

\section*{Acknowledgments}
We would like to thank and acknowledge the research engineers at Queen's University in the Laboratory for Percutaneous Surgery (Perk Lab) Dr. Tamas Ungi, Dr. Andras Lasso and Kyle Sunderland for their support during the development of SlicerROS2.

\vfill

\end{document}